\documentclass[letterpaper, 10 pt, conference]{ieeeconf}  

\IEEEoverridecommandlockouts                              

\usepackage{graphicx}
\usepackage{amsmath}
\usepackage{algorithm}
\usepackage{amssymb}
\usepackage{booktabs}

\usepackage{bigstrut}
\usepackage{bbm}
\usepackage{bbding}
\usepackage{bbold}
\usepackage{mathtools}

\usepackage{caption}
\usepackage{subcaption}
\usepackage{multirow}
\usepackage{xcolor}
\usepackage{hyperref}

\usepackage{amsthm}
\usepackage{cite}
\usepackage{hyperref}

\usepackage{fancyhdr}

\newtheorem{definition}{Definition}

\newcommand{\rebuttal}[1]{{\textcolor{black}{#1}}}

\definecolor{MyDarkRed}{rgb}{0.70980392, 0.09019608,0.0}
\definecolor{MyLightBlue}{rgb}{0.34117647, 0.75686275, 1.}
\definecolor{MyLightGreen}{rgb}{0.5372549 , 0.98431373, 0.30588235}

\title{\Large \bf
Benchmarking Reinforcement Learning Techniques for Autonomous Navigation
}

\author{Zifan Xu$^{1}$, Bo Liu$^{1}$, Xuesu Xiao$^{2, 3}$, Anirudh Nair$^{1}$, and Peter Stone$^{1, 4}$ 
\thanks{\scriptsize
        $^{1}$Department of Computer Science,
        University of Texas at Austin $^{2}$Department of Computer Science, George Mason University $^{3}$Everyday Robots
        $^{4}$Sony AI.
        This work has taken place in the Learning Agents Research Group (LARG) at UT Austin. LARG research is supported in part by NSF (CPS-1739964, IIS-1724157, NRI-1925082), ONR (N00014-18-2243), FLI
(RFP2-000), ARO (W911NF-19-2-0333), DARPA, Lockheed Martin, GM, and Bosch. Peter Stone serves as the Executive Director of Sony AI America and receives financial compensation for this work. The terms of this arrangement have been reviewed and approved by the University of Texas at Austin in accordance with its policy on objectivity in research.
}%
}

\begin{document}
\thispagestyle{fancy}

\maketitle
\thispagestyle{fancy}
\pagestyle{empty}

\begin{abstract}

Deep reinforcement learning (RL) has brought many successes for autonomous robot navigation. However, there still exists important limitations that prevent \emph{real-world} use of RL-based navigation systems.
    For example, most learning approaches lack safety guarantees;
and learned navigation systems may not generalize well to unseen environments.
Despite a variety of recent learning techniques to tackle these challenges in general, a lack of an open-source benchmark and reproducible learning methods specifically for autonomous navigation makes it difficult for roboticists to choose what learning methods to use for their mobile robots and for learning researchers to identify current shortcomings of general learning methods for autonomous navigation.
In this paper, we identify four major desiderata of applying deep RL approaches for autonomous navigation: 
(D1) \textit{reasoning under uncertainty},
(D2) \textit{safety},
(D3) \textit{learning from limited trial-and-error data}, and 
(D4) \textit{generalization to diverse and novel environments}. 
Then, we explore four major classes of learning techniques with the purpose of achieving one or more of the four desiderata:
\textit{memory-based neural network architectures }(D1), \textit{safe RL} (D2), \textit{model-based RL} (D2, D3), and \textit{domain randomization} (D4). By deploying these learning techniques in a new open-source large-scale navigation benchmark and real-world environments, we perform a comprehensive study aimed at establishing to what extent can these techniques achieve these desiderata for RL-based navigation systems.

\end{abstract}

\section{Introduction}
\label{sec::intro}

\begin{figure*}
    \centering
    \includegraphics[width=0.9\textwidth]{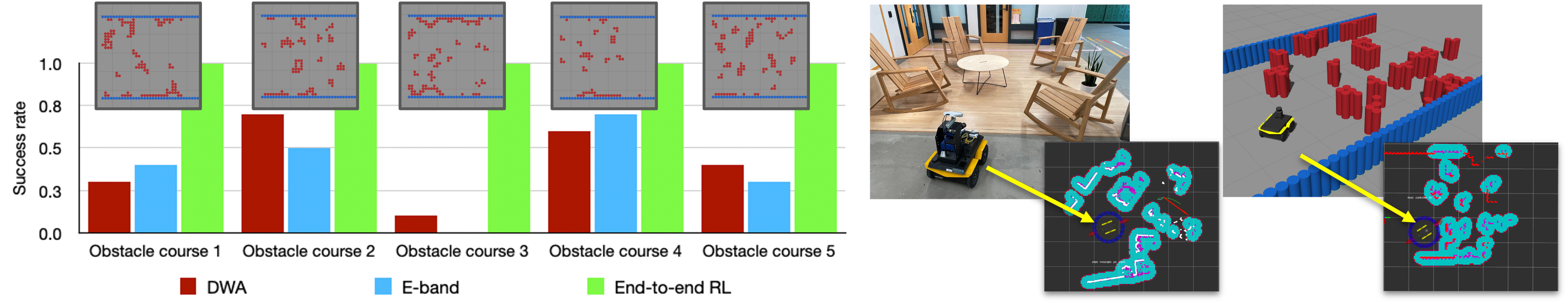}
    \caption{\textbf{Left:} Success rate of two classical navigation systems, DWA \cite{fox1997dynamic} (\textcolor{MyDarkRed}{red}) and E-band \cite{quinlan1993elastic} (\textcolor{MyLightBlue}{blue}), and vanilla end-to-end RL-based (\textcolor{MyLightGreen}{green}) navigation systems (individually trained) in five randomly generated difficult obstacle courses. The insets at the top show top-down views of the five obstacle courses. \textbf{Right:} Navigation environments in the real world (left) and the proposed benchmark (right) are similar to the robot perception system (e.g., white/red laser scans and cyan/purple costmaps).}
    \label{fig:compare}
\vspace{-15pt}
\end{figure*}

Autonomous robot navigation,  i.e., moving a robot from one point to another without colliding with any obstacle, has been studied by the robotics community for decades. 
Classical navigation systems \cite{quinlan1993elastic, fox1997dynamic} can successfully solve such navigation problem in many real-world scenarios, e.g., handling noisy, partially observable sensory input but still providing verifiable collision-free safety guarantees. However, these systems require extensive engineering effort, and can still be brittle in challenging scenarios, e.g., in highly constrained environments. This is reflected by a recent competition (The BARN Challenge~\cite{xiao2022autonomous}) held in ICRA 2022, which suggests that even experienced roboticists tend to underestimate how difficult navigation scenarios are for real robots.
Recently, data-driven approaches have also been used to tackle the navigation problem \cite{xiao2020motion} thanks to advances in the machine learning community. In particular, Reinforcement Learning (RL), i.e., learning from self-supervised trial-and-error data, has achieved tremendous progress on multiple fronts, including safety \cite{chow2019lyapunov, thomas2022safe, rodriguez2011lyapunov}, generalizability \cite{Cobbe2019QuantifyingGI, cobbe2019procgen, Justesen2018IlluminatingGI, tobin2017domain}, sample efficiency \cite{10.1145/122344.122377, nagabandi2018neural}, and addressing temporal data \cite{Hausknecht2015DeepRQ, Wierstra2007SolvingDM, chua2018deep}. For the problem of navigation, learned navigation systems from RL \cite{chiang2019learning} have the potential to relieve roboticists from extensive engineering efforts \cite{xiao2020appld, wang2021appli, wang2021apple, xu2021applr, xiao2021appl} spent on developing and fine-tuning classical systems. Moreover, a simple case study conducted in five randomly generated obstacle courses where classical navigation systems often fail shows that a RL-based navigation has the potential to achieve superior behaviors in terms of successful collision avoidance and goal reaching (Fig. \ref{fig:compare} left). 

Despite such promising advantages, learning-based navigation systems are far from finding their way into real-world robotics use cases, which currently still rely heavily on their classical counterparts. Such reluctance in adopting learning-based systems in the real world stems from a series of fundamental limitations of learning methods, e.g., lack of safety, explainability, and generalizability. 
To make things even worse, a lack of well-established comparison metrics and reproducible learning methods further obfuscates the effects of different learning approaches on navigation across both the robotics and learning communities, making it difficult to assess the state of the art and therefore to adopt learned navigation systems in the real world.

To facilitate research in developing RL-based navigation systems with the goal of deploying them in real-world scenarios, we introduce a new open-source large-scale navigation benchmark with a variety of challenging, highly constrained obstacle courses to evaluate different learning approaches, along with the implementation of several state-of-the-art RL algorithms. The obstacle courses resemble highly-constraint real-world navigation environments (Fig. \ref{fig:compare} right), and present major challenges to existing classical navigation systems, while RL-based navigation systems have the potential to perform well in them (Fig. \ref{fig:compare} left). 

We identify four major desiderata that ought to be fulfilled by any learning-based system that is to be deployed: (D1) \textit{reasoning under uncertainty of partially observed sensory inputs}, (D2) \textit{safety}, (D3) \textit{learning from limited trial-and-error data}, and (D4) \textit{generalization to diverse and novel environments}. By deploying four major classes of learning techniques: \textit{memory-based neural network architectures}, \textit{safe RL}, \textit{model-based RL}, and \textit{domain randomization}, 
we perform extensive experiments and empirically compare a large range of RL-based methods based on the degree to which they achieve each of these desiderata. 
Moreover, by deploying six selected navigation systems in three qualitatively different real-world navigation environments, we investigate to what degree the conclusions drawn from the benchmark can be applied to the real world. Supplementary videos and material for this work are available on the \href{https://cs.gmu.edu/~xiao/Research/RLNavBenchmark/}{project webpage}.\footnote{\href{https://cs.gmu.edu/~xiao/Research/RLNavBenchmark/}{https://cs.gmu.edu/~xiao/Research/RLNavBenchmark/}}

\section{Desiderata for Learning-based Navigation}
\label{sec::challenge}
In this section, we introduce four desiderata for learning-based autonomous navigation systems and briefly discuss the learning techniques as their corresponding solutions.

\textbf{(D1) reasoning under uncertainty of partially observed sensory inputs.} Autonomous navigation without explicit mapping and localization is usually formalized as a Partially Observable Markov Decision Process (POMDP), where the agent produces the motion of the robot only based on limited sensory inputs that are usually not sufficient to recover the full state of the navigation environment. Most RL approaches solve POMDPs by maintaining a history of past observations and actions \cite{Hausknecht2015DeepRQ, Wierstra2007SolvingDM}. Then, neural network architectures like Recurrent Neural Networks (RNNs) that process sequential data are employed to encode history and address partial observability. In this study, we investigate various design choices of history-dependent architectures.

\textbf{(D2) safety.} Even though in some cases deep RL methods achieve comparable performance to classical navigation, they still suffer from poor explainability and do not guarantee collision-free navigation. The lack of safety guarantee is a major challenge preventing RL-based navigation from being used in the real world. Prior works have addressed this challenge by formalizing the navigation as a multi-objective problem that treats collision avoidance as a separate objective from reaching the goal and solving it with Lagrangian or Lyapunov-based methods \cite{chow2019lyapunov}. 
For simplicity, we only explore Lagrangian method and investigate whether explicitly treat safety as a separate objective leads to safer and smoother learned navigation behavior.

\textbf{(D3) learning from limited trial-and-error data.} Although deep RL approaches can alleviate roboticists from extensive engineering effort, a large amount of data is still required to train a typical deep RL agent. However, autonomous navigation data is usually expensive to collect in the real world. Therefore, data collection is usually conducted in simulation, e.g., in the Robot Operating System (ROS) Gazebo simulator, which provides an easy interface with real-world robots. 
However, simulating a full navigation stack from perception to actuation is more computationally expensive compared to other RL domains, e.g., MuJuCo or Atari games \cite{6386109, bellemare13arcade}, which presents a high requirement for sample efficiency. Most prior works have used off-policy RL algorithms to improve sample efficiency with experience replay \cite{Chiang2019LearningNB, Wahid2019LongRN}. In addition, model-based RL methods can explicitly improve sample efficiency, and are widely used in robot control problems. In this study, we compare two common classes of model-based RL method \cite{10.1145/122344.122377, nagabandi2018neural} combined with an off-policy RL algorithm, and empirically study to what extent model-based approaches improve sample efficiency when provided with different amounts of data. 

\textbf{(D4) generalization to diverse and novel environments.} The ultimate goal of deep RL approaches for autonomous navigation is to learn a generalizable policy for all kinds of navigation environments in the real world. 
A simple strategy is to train the agent in as many diverse navigation environments as possible or domain randomization, but it is unclear what is the necessary amount of training environments to efficiently achieve good generalization. Utilizing the large-scale navigation benchmark proposed in this paper, we empirically study the dependence of generalization on the number of training environments. 
\section{Navigation Benchmark}
\label{sec::benchmark}
This section details the proposed navigation benchmark for RL-based navigation systems, which aims to provide a unified and comprehensive testbed for future autonomous navigation research. \rebuttal{First, Sec. \ref{sec::related} discusses the difference between the proposed benchmark and existing navigation benchmarks.} In Sec. \ref{subsec::nav_task} and \ref{subsec::pd}, the navigation task is formally defined and formulated as a POMDP. More detailed background of MDP and POMDP can be found on the \href{https://cs.gmu.edu/~xiao/Research/RLNavBenchmark/}{project webpage}. Finally, Sec. \ref{subsec::nav_env} introduces 
simulated and real-world environments that benchmark different aspects of navigation performance.

\subsection{Existing Navigation Benchmarks}
\label{sec::related}
\rebuttal{
Our proposed benchmark differs from existing benchmarks in three aspects: (1) \textbf{high-fidelity physics}: the navigation tasks are simulated by Gazebo~\cite{koenig2004design}, which is based on realistic physical dynamics and therefore tests motion planners that directly produce low-level motion commands, i.e., linear and angular velocies, in contrast to high-level instructions such as turn left, turn right, move forward \cite{zhu2017target, harries2019mazeexplorer}. In other words, we focus on ``how to navigate" (motion planning), instead of ``where to navigate" (path planning); (2) \textbf{ROS integration}: our benchmark is based on ROS \cite{ros}, which allows seamless transfer of a navigation method developed and benchmarked in simulation directly onto a physical robot with little (if any) effort; and (3) \textbf{collision-free navigation}: the benchmark includes both static and dynamic environments, and requires \emph{collision-free} navigation, whereas other benchmarks assume that either collisions are possible \cite{harries2019mazeexplorer} or collision-avoidance will be addressed by other low-level controllers out of the scope of the benchmark \cite{zhu2017target}. A special case is the photo-realistic interactive Gibson benchmark by Xia et. al.~\cite{xia2020interactive}, which intentionally allows physical interaction with objects (e.g., pushing) and therefore pose no challenges to the collision-avoidance system.}

\subsection{Navigation Problem Definition}
\label{subsec::nav_task}
\begin{definition}[Robot Navigation Problem]
Situated within a navigation environment $e$ which includes information of all the obstacle locations at any time $t$, a start location $(x_i, y_i)$, a start orientation $\theta_i$, and a goal location $(x_g, y_g)$, the navigation problem $\mathcal{T}_e$ is to maximize the probability $p$ of a mobile robot reaching the goal location from the start location and orientation under a constraint on the number of collisions with any obstacle $C < 1$ and a time limit $t < T_{max}$.
\label{def::navigation_formal}
\end{definition}
 A navigation problem can be formally defined as above. Given the current location $(x_t, y_t)$, the robot is considered to have reached the goal location if and only if its distance to the goal location is smaller than a threshold, $d_t < d_s$, where $d_t$ is the Euclidean distance between $(x_t, y_t)$ and $(x_g, y_g)$, and $d_s$ is a constant threshold.

\subsection{POMDP Formulation}
\label{subsec::pd}
A navigation task $\mathcal{T}_e$ can be formulated as a POMDP conditioned on a navigation environment $e$, which can be represented by a 7-tuple $(S_e, A_e, O_e, T_e, \gamma_e, R_e, Z_e)$ . In this POMDP, the state $s_t \in S_e$ is a 5-tuple $(x_t, y_t, \theta_t, c_t, e)$ with $x_t, y_t, \theta_t$ the two-dimensional coordinates and the orientation of the robot at time step $t$, $c_t$ a binary indicator of whether a collision has occurred since the last time step $t-1$, and $e$ the navigation environment. The action $a_t=(v_t, \omega_t) \in A_e$ is a two-dimensional continuous vector that encodes the robot's linear and angular velocity. The observation $o_t = (\chi_t, \bar{x_t}, \bar{y_t}) \in O_e$ is a 3-tuple composed of the sensory input $\chi_t$ from LiDAR scans and the relative goal position $(\bar{x_t}, \bar{y_t})$ in the robot frame. The observation  model $Z: S \rightarrow O$ maps the state to the observation. The reward function for this POMDP is defined as follows:
\begin{equation}
R_e(s_t, a_t) = + b_f \cdot \mathbb{1} (d_{t} < d_s) + b_p \cdot (d_{t-1} - d_{t}) - b_c \cdot c_t,
\label{eqn::reward}
\end{equation}
where $\mathbb{1} (d_{t} < d_s)$ is the indicator  function of reaching the goal location, $d_t$ is the Euclidean distance to the goal location, and $b_f$, $b_p$, $b_c$ are the coefficient constants. In this reward function, the first term is the true reward function that assigns a positive constant $b_f$  for the success of an agent, which matches with the objective of the navigation task in Definition \ref{def::navigation_formal}. The second and third terms are auxiliary rewards that facilitate the training by encouraging local progress and penalizing collisions. 

We perform a grid search over different values of the coefficients in this reward function, and the result shows that the auxiliary reward term $(d_{t-1} - d_t)$ is necessary for successful training,
and a much smaller coefficient $b_p$ relative to $b_f$ can lead to a better asymptotic performance. The agent can learn without the penalty reward for collision ($b_c = 0$), but a moderate value of $b_c$ can improve the asymptotic performance and speed up training. For all the experiments in this paper, we fix the coefficients as $b_f=20$, $b_p=1$ and $b_c=4$.

In our experiments, the RL algorithm solves
a multi-task RL problem where the tasks are randomly sampled from a task distribution $\mathcal{T}_e \sim p(\mathcal{T}_e)$. Here the task distribution $p(\mathcal{T}_e):=U(\{e_i\}_{i=1}^{N})$ is a uniform distribution on a set of $N$ navigation environments $\{e_i\}_{i=1}^{N}$. The overall objective of this multi-task RL problem is to find an optimal policy $\pi^* = \max_\pi \mathbb{E}_{\mathcal{T}_e \sim p(\mathcal{T}_e), \tau_t \sim \pi} \big[\sum_{t=0}^\infty \gamma^t R_e(s_t, a_t)\big]$.

\subsection{Navigation Environments}
\label{subsec::nav_env}

\begin{figure}
    \centering
    \includegraphics[width= \linewidth]{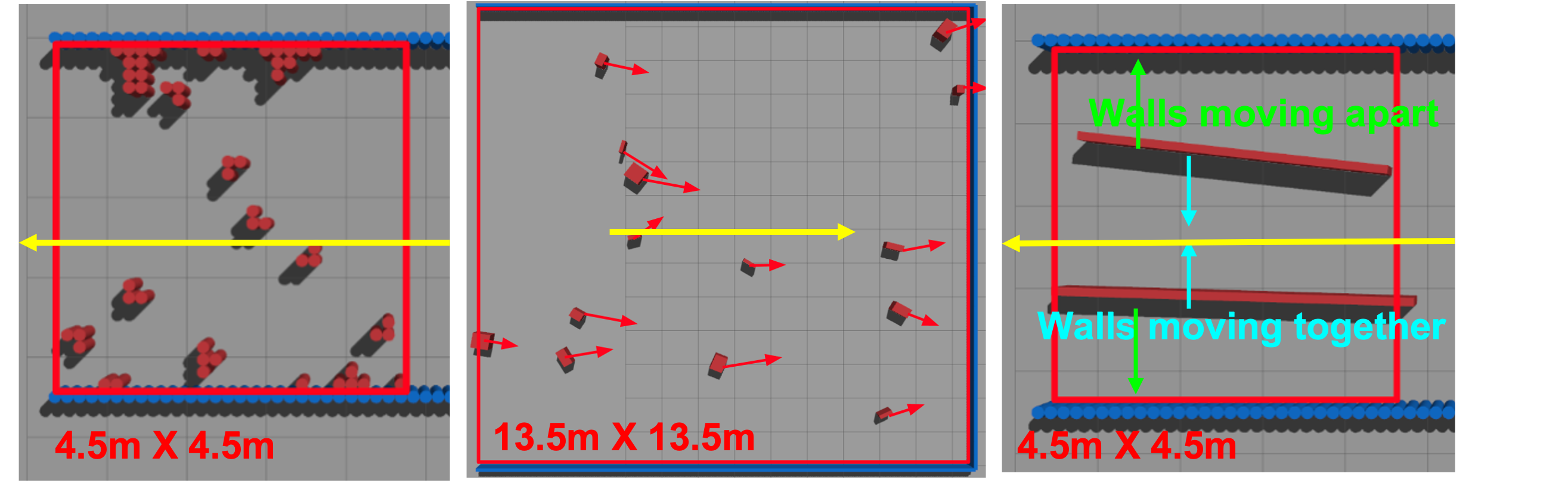}
    \centering
    \caption{Three types of navigation environments: \texttt{static} (left), \texttt{dynamic box} (middle), and \texttt{dynamic-wall} (right). The red squares mark the obstacle fields, and the yellow arrows mark the direction of navigation. In \texttt{dynamic-wall}, the green (blue) arrows indicate the case when the two walls are moving apart (together). In \texttt{dynamic box}, the red arrows indicate the velocities of obstacles.}
    \label{fig:nav_env}
\vspace{-15pt}
\end{figure}

The navigation is performed by a ClearPath Jackal differential-drive ground robot in simulated by the Gazebo simulator. More details of the robot and simulation can be found on the \href{https://cs.gmu.edu/~xiao/Research/RLNavBenchmark/RLNavBenchmark.html}{project webpage}. Each environment in this benchmark will have a navigation system navigating the robot through a 10m navigation path that passes through a highly constrained obstacle course. Walls are placed at three edges of a square so that passing through the obstacle field is the only path to the goal location (see Fig. \ref{fig:nav_env}). The benchmark includes 300 \texttt{static} environments, 100 \texttt{dynamic-box} environments, and 100 \texttt{dynamic-wall} environments. The \texttt{static} environments contains a diverse set of obstacle course covering a large range of difficulty levels from easy to hard. A \texttt{dynamic-box} environment has small boxes with random shapes and velocities to test the system's immediate reactions to small moving obstacles. A \texttt{dynamic-wall} has two walls moving oppositely that requires the system to make a longer-term decision of whether to pass or wait. The detailed procedures of generating these environments can be found on the \href{https://cs.gmu.edu/~xiao/Research/RLNavBenchmark/}{project webpage}. We randomly select 50 environments from each type as the test sets, which are denoted as \texttt{static-test}, \texttt{dynamic-box-test}, and \texttt{dynamic-wall-test}. The remaining environments are denoted as \texttt{static-train}, \texttt{dynamic-box-train}, and \texttt{dynamic-wall-train} respectively. To study the effect of randomization, \texttt{static-train} is further separated as \texttt{static-train-5}, \texttt{static-train-10}, \texttt{static-train-50}, \texttt{static-train-100}, and \texttt{static-train-250} by randomly sampling 5, 10, 50, 100, and all 250 environments from \texttt{static-train}.

\begin{figure}
    \centering
    \includegraphics[width=0.75 \columnwidth]{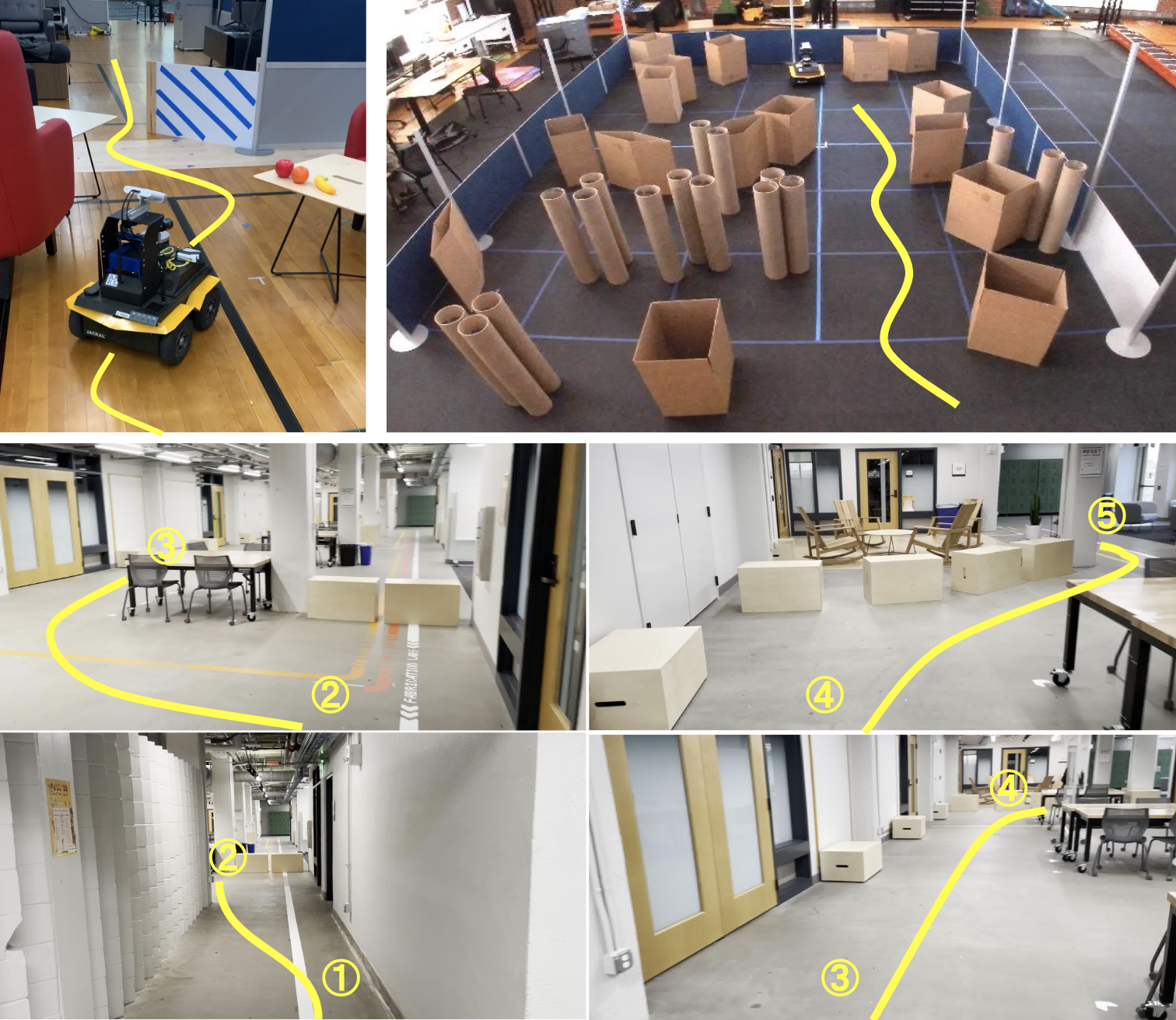}
    \caption{Real-world benchmark-like (top-right), in-door highly-constrained (top-left), and large-scale (bottom) environments. The yellow curves mark the paths of navigation.} 
    \label{fig:real_world}
\vspace{-15pt}
\end{figure}

\label{sec:physical_exp_speci}
To test the sim-to-real transferability of the policies learning with different techniques, the navigation systems are deployed in three qualitatively different static navigation environments including a benchmark-like environment (Fig. \ref{fig:real_world} left), an indoor highly-constrained environment (Fig. \ref{fig:real_world} right), and a large-scale environment of 30 meters in length. We denote them as \texttt{real-world-1}, \texttt{real-world-2}, and \texttt{real-world-3} respectively.

\section{Experiments}
In this section, we present experimental results of each studied technique to achieve the proposed desiderata in Sec. \ref{sec::challenge}. We implement distributed training pipelines (similar to \cite{xu2021applr}) of different RL algorithms including TD3\cite{fujimoto2018addressing}, SAC\cite{haarnoja2018soft}, and DDPG\cite{lillicrap2015continuous}. They perform similarly in the study of different neural network architectures. For simplicity, all the experiments mentioned in this section use TD3 combined with the corresponding techniques, and all the data points are averaged over three independent runs. 

\begin{table*}[!htb]
\centering
\resizebox{0.85\linewidth}{!}{%
\begin{tabular}{lccccccccc}
\toprule 
\multirow{2}{*}{\rebuttal{Success rate (\%) ($\uparrow$)} }   & \multicolumn{3}{c}{\texttt{static env.}}                                          & \multicolumn{3}{c}{\texttt{dynamic-box env.}}                                          & \multicolumn{3}{c}{\texttt{dynamic-wall env.}}                                           \\ 
 \cmidrule(lr){2-4} \cmidrule(lr){5-7} \cmidrule(lr){8-10}
  & $H=1$                       & $H=4$                       & $H=8$         & $H=1$                       & $H=4$                       & $H=8$          & $H=1$          & $H=4$                       & $H=8$                      \\  \midrule 

MLP            & $65 \pm 4$ & $57 \pm 7$              & $42 \pm 2$ & $50 \pm 5$ & $35 \pm 2$              & $46 \pm 3$ & $67 \pm 7$ & $72 \pm 1$              & $69 \pm 4$              \\
GRU            & -                       & $51 \pm 2$              & $43 \pm 4$ & -                       & $48 \pm 4$              & $45 \pm 1$ & -          & $\boldsymbol{82 \pm 4}$ & $78 \pm 5$ \\
CNN            & -                       & $55 \pm 4$              & $45 \pm 5$ & -                       & $42 \pm 5$              & $40 \pm 1$ & -          & $63 \pm 3$              & $43 \pm 3$              \\
Transformer    & -                       & $\boldsymbol{68 \pm 2}$ & $46 \pm 3$ & -                       & $\boldsymbol{52 \pm 1}$ & $44 \pm 4$ & -          & $33 \pm 28$             & $15 \pm 13$             \\ 
 \bottomrule
\end{tabular}%
}
\caption{(D1) Success rate (\%) \rebuttal{($\uparrow$)} of policies trained with different neural network architectures and history lengths. \rebuttal{$H$ is the history length of the memory. Bold font indicates the best success rate for each type of environment.}} 
\label{tab::arch}

\centering
\resizebox{0.75\linewidth}{!}{%
\begin{tabular}{lccccc}
\toprule
Methods & Baseline (model-free) & Lagrangian method & MPC (model-based) & DWA & \rebuttal{TEB}\\ \midrule
Success rate (\%) {($\uparrow$)} & $65 \pm 4$ & $74 \pm 2$ & $70 \pm 3$ & $\boldsymbol{82}$ & \rebuttal{70}\\ \midrule
Survival time (s) {($\uparrow$)} & $8.0 \pm 1.5$ & $16.2 \pm 2.5$ & $55.7 \pm 4.9$ & $\boldsymbol{62.7}$ & \rebuttal{26.9}\\
Traversal time (s) {($\downarrow$)} & $\boldsymbol{7.5 \pm 0.3}$ & $8.6 \pm 0.2$ & $24.7 \pm 2.0$ & $35.6$ & \rebuttal{26.9}\\
\bottomrule
\end{tabular}
}
\caption{(D2) Success rate \rebuttal{($\uparrow$)}, survival time \rebuttal{($\uparrow$)}, and traversal time \rebuttal{($\downarrow$)} of policies trained with Lagrangian method, MPC with probabilistic transition model, and DWA. \rebuttal{The bold font indicates the best number achieved for each type of metric.}}
\label{tab::saferl1}
\vspace{-15pt}
\end{table*}

\subsection{Memory-based Neural Network Architectures (D1)}

To benchmark the performance of different neural network (NN) architectures, deep RL policies represented by architectures of Multilayer Perceptron (MLP), One-dimensional Convolutional Neural Network (CNN), Gated Recurrent Units (GRU), and Transformer with history length of 4 and 8 are trained in \texttt{static-train-50}, and the two types of dynamic environments \texttt{dynamic-box-train} and \texttt{dynamic-wall-train} from Sec. \ref{subsec::nav_env}. After training, the policies are tested in their corresponding test sets. In addition, MLP with history length of one is added as a memory-less baseline. Table \ref{tab::arch} shows the success rates of policies with different architectures and history lengths evaluated in \texttt{static-test} (
left), \texttt{dynamic-wall-test} (middle) and \texttt{dynamic-box-test} respectively.

\textbf{Memory-based NNs only marginally improve navigation performance in static environments.} In Table \ref{tab::arch}, the policy represented by Transformer with a history length of 4 shows the best success rate of $68 \%$, with a slightly worse success rate of $65 \%$ achieved by the baseline MLP. Additionally, a monotonic decrease in success rate with increasing history length is observed in each tested NN architecture. For example, a $32\%$ drop in the success rate of Transformer is shown by increasing the history length from 4 to 8. One possible explanation is that, if only few past observations are useful to make the decision,  including more history will make it more difficult to learn a generalized policy in this very diverse training set. 

\textbf{Memory is essential when possible catastrophic failures will happen by making the wrong long-term decisions.} Memory usually matters for dynamic environments when a single time frame is not sufficient to estimate the motion of obstacles. Surprisingly, in \texttt{dynamic-box} where the dynamic obstacles are completely random, the memory-based NN architectures do not outperform the memory-less baseline. On the other hand, in \texttt{dynamic-wall} with a manually designed dynamic challenge, the best success rate of $82 \%$ is observed in GRU with a history length of 4, which improves about $15 \%$ over the non-memory baseline. During our deployment of the policies, we observe that, in \texttt{dynamic-box} even though the memory-less agent does not estimate the motion and adjust its plan in advance, it tends to perform safely and avoids the obstacles when they get close enough. This simple strategy works surprisingly well and achieves similar success rate as the memory-based policies. However, this strategy does not work in the manually designed dynamic challenges like \texttt{dynamic-wall} where the agent has to estimate the motion of the obstacles to pass safely.

\subsection{Safe RL (D2)}

To investigate to what extent safe RL methods can help to improve safety, a TD3 agent with the Lagrangian-based safe RL method is trained in \texttt{static-train-50}, and then tested in \texttt{static-test}. The policy is represented by a MLP with its input containing only one history length. Table \ref{tab::saferl1} shows the success rate, average survival time, and average traversal time of the safe RL agent trained with Lagrangian method and a baseline MLP agent tested in \texttt{static-test}. We define survival time as the time cost of an unsuccessful episode (collision or exceeding a time limit of 80s). Traversal time, instead, is the time cost of a successful episode. With the same level of success rate, a longer survival time means that the agent tends to, at least, avoid collisions if it cannot succeed. To compare the safe RL method with classical navigation systems which are believed to have better safety, we also add evaluation metrics from a classical navigation stack with the Dynamic Window Approach (DWA)  \cite{fox1997dynamic} local planner.

\textbf{Lagrangian method reduces the gap between training and test environments.} When deployed in the training environments, both the baseline MLP and the safe RL method achieves about $80 \%$ success rate. However, in the test environments, the Lagrangian method has a better success rate of $74\%$ compare to $65\%$ by the baseline MLP. We hypothesize that the safety constraint applied by the safe RL methods forms a way of regularization, and therefore, improves the generalization to unseen environments. 

\textbf{Lagrangian method increases the average survival time in failed episodes.} As expected, the Lagrangian method increases the average survival time by $8.2s$ compared to the baseline MLP at a cost of $1.1s$ longer average traversal time. However, such improved safety are still worse than the classical navigation systems given the best survival time of $88.6s$ achieved by DWA. 

\subsection{Model-based RL (D2 and D3)}
 To explore how the model-based approaches help with the autonomous navigation tasks, we implement Dyna-style, MPC, and MBPO, and evaluate the methods in static environments. The transition models are either represented by a deterministic NN or a probabilistic NN that predicts the mean and variance of the next state. During the training in \texttt{static-train-50}, the policies are saved when 100k, 500k and 2000k transition samples are collected, then tested in \texttt{static-test}. The success rates of these policies are reported in Table \ref{tab::model_based}.
 
 \textbf{Model-based methods do not improve sample efficiency.}
 As shown in the second and third columns in Table \ref{tab::model_based}, better success rates of $13\%$ and $58\%$ are achieved by the baseline MLP method provided by limited 100k and 500k transition samples respectively. In addition, Higher success rates at 500k transition samples are observed in probabilistic models compared to their deterministic counterparts, which indicates a more efficient learning with probabilistic transition models. Notice that MBPO exploits more heavily on the model compared to the Dyna-style method, which leads to much worse asymptotic performance (about 20\% success rate in the end).
 
 \textbf{Model-based methods with probabilistic dynamic models improve the asymptotic performance.} In the last column of Table \ref{tab::model_based}, both Dyna-style and MPC with probabilistic dynamic models achieve slightly better success rates of $70 \%$ compared to $65 \%$ in the baseline MLP method when sufficient transition samples of 2000k are given to the learning agent.  
 
 \textbf{The MPC policy performs conservatively when deployed in unseen test environments and shows a better safety performance.} The safety performances of MPC policies with probabilistic dynamic models are also tested (see Table \ref{tab::saferl1}). We observe that the agents with MPC policies navigate very conservatively with an average traversal time of $24.7s$, which is about two times more than the MLP baseline. In the meantime, MPC policies achieve improved safety with the best survival time of $55.7s$ among the RL-based methods. 
 
 \begin{table*}
\centering
\begin{minipage}[b]{0.28\linewidth}
\centering
\includegraphics[width=0.9\linewidth]{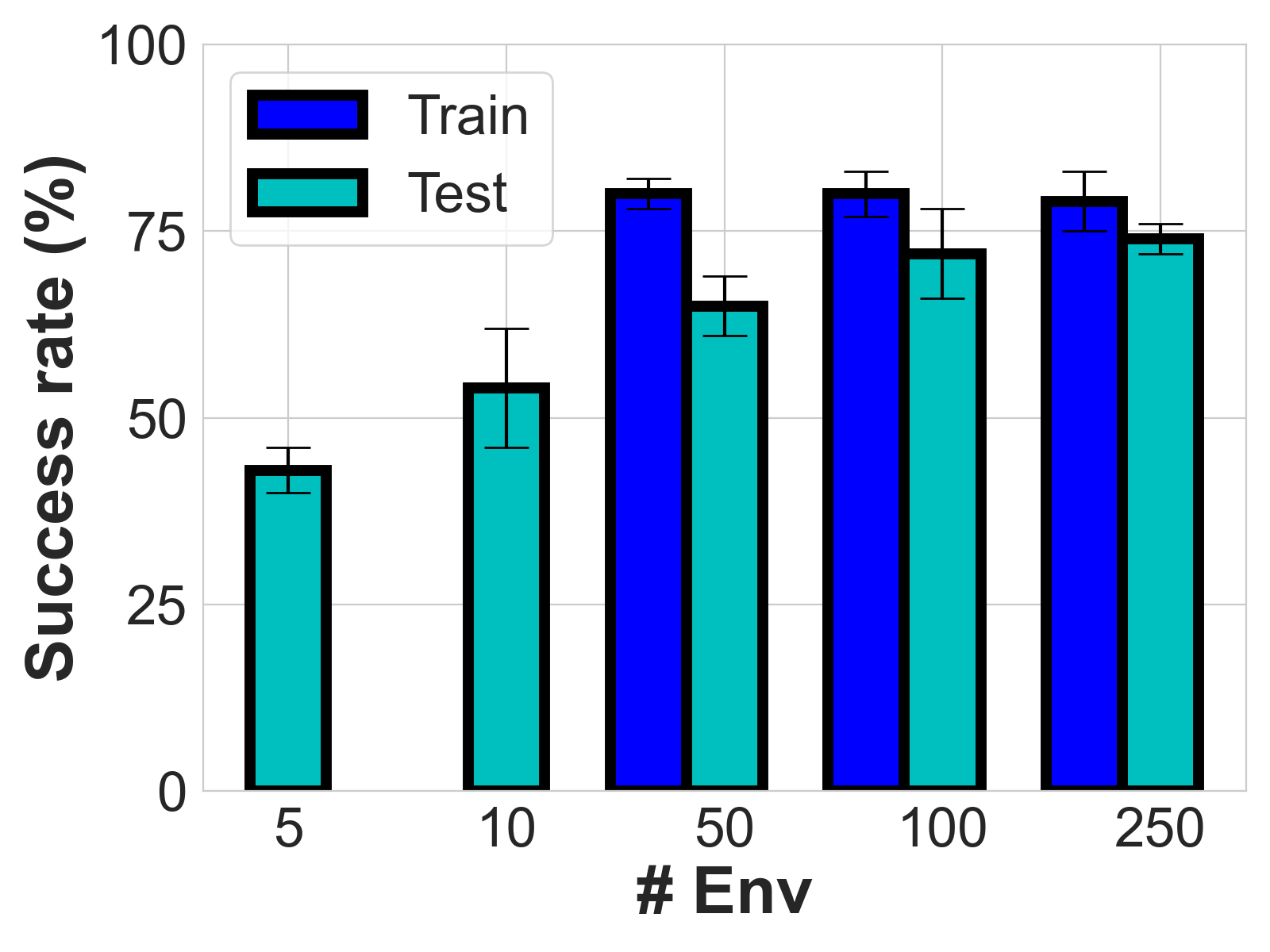}
\captionof{figure}{(D4) Success rate (\%) of policies trained with different number of training environments.}
\label{fig:gen}
\end{minipage}\hfill
\begin{minipage}[b]{0.69\linewidth}
\resizebox{\linewidth}{!}{%
\begin{tabular}{l|cc|ccc}
\toprule
            & \multirow{2}{*}{H} & \multirow{2}{*}{\# envs} & \texttt{real-world-1}         & \texttt{real-world-2}          & \texttt{real-world-3}          \\
            &  & & \multicolumn{3}{c}{\rebuttal{traversal time ($\downarrow$) ~~~(\# successful trials ($\uparrow$) / total \# trials)       }} \\ 
            
            \midrule
MLP         & 1              & 50                       & $6.9~(1/3)$         & $10.6~(1/3)$         & $\text{N}~(0/3)$             \\
MLP         & 1              & 250                      & $4.6 \pm 0.8~(3/3)$ & $\boldsymbol{6.6 \pm 0.6~(3/3)}$  & $\boldsymbol{22.6 \pm 0.5~(3/3)}$ \\
Transformer & 4              & 50                       & $6.1 \pm 0.4~(3/3)$ & $6.1 \pm 0.1~(2/3)$  & $20.5 \pm 2~(2/3)$   \\
Lagrangian      & 1              & 50                       & $\boldsymbol{4.4 \pm 0.6~(3/3)}$  & $7.1 \pm 0.1~(2/3)$  & $26.2~(1/3)$          \\
MPC         & 1              & 50                       & $13.2 \pm 0.7~(3/3)$ & $24.8 \pm 3.7~(3/3)$ & N $(0/3)$            \\
DWA         & -              & -                        & $16.2 \pm 0.7~(3/3)$ & $35.2 \pm 8.2~(2/3)$ & $66.9 \pm 0.6~(3/3)$ \\ \bottomrule
\end{tabular}%
}
\caption{Physical experiments. The table shows the traversal time (s) \rebuttal{($\downarrow$)} and the number of successful trials \rebuttal{($\uparrow$)} of 5 RL-based navigation systems and a classical navigation system (DWA) evaluated in three real-world environments. \rebuttal{The bold font indicates the best traversal time when all three trials are successful.} \label{tab:physical_exp}}
\end{minipage}\hfill
\vspace{-20pt}
\end{table*}
 
\vspace{-5pt}

 \subsection{Domain Randomization (D4)}
 To explore how model generalization depends on the degree of randomness in the training environments, baseline MLP policies with one history length are trained in the environment sets with 5, 10, 50, 100, and 250 training environments.
 The trained policies are tested in the same \texttt{static-test}. To investigate the performance gap between training and test, the policies trained with 50, 100, and 250 environments are also tested on \texttt{static-train-50}, which is part of their training sets. Fig \ref{fig:gen} shows the success rate of policies trained with different number of training environments.
 
 \textbf{The generalization to unseen environments improves with increasing number of training environments.} As shown in Fig.~\ref{fig:gen}, the performances on the unseen test environments monotonously increase from 43\% to 74\% with the number of training environments increasing from 5 to 250. Moreover, the gaps between training and test environments gradually shrink by adding more training environments provided by that the polices are robust enough to maintain similar performances of about 80\% on the training environments.

\begin{table}[!htb]
\centering
\resizebox{0.9\linewidth}{!}{%
\centering
\begin{tabular}{lccccc}
\toprule
  Transition samples & 100k & 500k & 2000k\\ \midrule
MLP & $\boldsymbol{13 \pm 7}$ & $\boldsymbol{58 \pm 2}$ & $65 \pm 4$\\
Dyna-style deterministic & $8 \pm 2$ & $30 \pm 10$ & $66 \pm 5$\\
MPC deterministic& $0 \pm 0$ & $21 \pm 10$ & $62 \pm 3$\\ 
Dyna-style probabilistic & $0 \pm 0$ & $48 \pm 4$ & $\boldsymbol{70 \pm 1}$\\
MPC probabilistic& $0 \pm 0$ & $45 \pm 4$ & $\boldsymbol{70 \pm 3}$\\
MBPO& $\rebuttal{0 \pm 0}$ & $\rebuttal{0 \pm 0}$ & $\rebuttal{21.9 \pm 3}$\\
\bottomrule
\end{tabular}%
}
\caption{(D3) Success rate (\%) \rebuttal{($\uparrow$)} of policies trained with different model-based methods and different number of transition samples. \rebuttal{The bold font indicates the best success rate for each number of transition samples.}}
\label{tab::model_based}
\vspace{-15pt}
\end{table}

 \subsection{Physical experiments}
To study the consistency of the above observations in simulation and the real world, we deploy one baseline MLP policy, one best policy for each studied desideratum, and one classical navigation system (DWA \cite{fox1997dynamic}) in the three real-world environments introduced in Sec. \ref{sec:physical_exp_speci}. Each deployment is repeated three times, and the average traversal time and the number of successful trials are reported in Table. \ref{tab:physical_exp}. Even though the best memory-based policy, transformer architecture with 4 history length, was only marginally better than the baseline MLP in simulation, in the real world it can navigate very smoothly and fails only once in \texttt{real-world-2} and \texttt{real-world-3}, while baseline MLP fails most of the trials in all the environments including the benchmark-like environment. One possible reason for this is that simulations are typically more predictable than the real world. Therefore, it is particularly important to use historical data in the real world to estimate the environment and current states of the robot. Similarly, MLP policy trained with 250 environments can successfully navigate in all the environments without any failures, while baseline MLP trained with 50 environments fails most of the trials. Safe RL improves the chances of success in all the environments and can navigate more safely by performing backups and small adjustments of robots' poses. Similar to the simulation, MPC navigates very conservatively and succeeds in all the trials in \texttt{real-world-1} and  \texttt{real-world-2}, but has much more difficulty generalizing to large-scale \texttt{real-world-3}.
\section{Conclusion}
\rebuttal{In this section, we discuss the conclusions we draw from these benchmark experiments. We organize these conclusions by the desiderata as follows:}

\textbf{(D1) reasoning under uncertainty of partially observed sensory inputs} does not obviously benefit from adding memory in simulated static environments and very random dynamic (\texttt{dynamic-box}) environments, but much more significant improvements were observed in the \emph{real world} and in more \emph{challenging dynamic environments} (\texttt{dynamic-wall}).

\textbf{(D2) safety} is improved by both safe RL and model-based MPC methods. However, classical navigation systems still achieve the best safety performance at a cost of very long traversal time. Whether RL-based navigation systems can achieve similar safety guarantees as classical navigation systems and whether safety can be improved without significantly sacrificing the traversal time are still open questions.

\textbf{(D3) the ability to learn from limited trial-and-error data} is not improved by the evaluated model-based methods.
Currently, we observe that model-based RL methods indeed improve sample-efficiency, but only when the number of imaginary rollouts from the learned model is large (e.g. $\geq 2000k$) and when they are sampled with randomness. We therefore hypothesize that the improvement comes from the robustness brought by learning on more data sampled from the learned model. Hence, this result motivates not only more accurate model learning for reducing the number of imaginary rollouts, but also theoretical understanding of how the model helps improve the robustness or even safety of navigation.

\textbf{(D4) the generalization to diverse and  novel environments} is improved by increasing the randomness of training environments. However, a noticeable gap of about $5\%$ between training and test environments is not eliminated by further increasing the number of training environments to 250. This reflects the limitation of simple \emph{domain randomization} to increase the generalization, which is, however, widely used by the community.

In summary, although the proposed benchmark is not intended to represent every real-world navigation scenario, it serves as a simple yet comprehensive testbed for RL-based navigation methods. We observed that for every desideratum, no method can achieve 100\% success rate on all \emph{training} environments. Even though we ensured that we have made sure that every environment is indeed individually solvable. This alone indicates that there exists an optimization and generalization challenge when we have a large number of training environments as in our proposed benchmark.

\newpage
\bibliographystyle{IEEEtran}
\bibliography{IEEEabrv,references}







\end{document}